\documentclass{esannV2}
\usepackage[dvips]{graphicx}
\usepackage{epstopdf}
\usepackage[latin1]{inputenc}
\usepackage{amssymb,amsmath,array}
\usepackage{hyperref}
\usepackage{multirow}

\usepackage{eso-pic}
\newcommand\AtPageUpperMyright[1]{\AtPageUpperLeft{%
		\put(\LenToUnit{0.5\paperwidth},\LenToUnit{-3cm}){%
			\parbox{1.45\textwidth}{\raggedleft\fontsize{9}{11}\selectfont #1}}%
}}%
\newcommand{\conf}[1]{%
	\AddToShipoutPictureBG*{%
		\AtPageUpperMyright{#1}
	}
}

\begin{document}
\title{GraN: An Efficient Gradient-Norm Based Detector for Adversarial and Misclassified Examples}

\author{Julia Lust$^{1,2}$ and Alexandru P. Condurache$^{1,2}$
%
%
\vspace{.3cm}\\
%
1- University of Lübeck - Institute for Signal Processing 
%
\vspace{.1cm}\\
2- Robert Bosch GmbH - Automated Driving \\
}


\maketitle
\conf{\textsf{ESANN 2020 proceedings, European Symposium on Artificial Neural Networks, Computational Intelligence\\
	and Machine Learning. Bruges (Belgium), 2-4 October 2020, i6doc.com publ., ISBN 978-2-87587-074-2.}
}
\begin{abstract}
Deep neural networks (DNNs) are vulnerable to adversarial examples and other data perturbations. Especially in safety critical applications of DNNs, it is therefore crucial to detect misclassified samples. The current state-of-the-art detection methods require either significantly more runtime or more parameters than the original network itself. This paper therefore proposes GraN, a time- and parameter-efficient method that is easily adaptable to any DNN. \\
GraN is based on the layer-wise norm of the DNN's gradient regarding the loss of the current input-output combination, which can be computed via backpropagation. GraN achieves state-of-the-art performance on numerous problem set-ups.
\end{abstract}

\section{Introduction}
\label{sec:intro}
Deep neural networks (DNNs) have proven themselves by achieving state-of-the-art performance in many application fields. Due to their good results, they are also used in safety-critical approaches, such as perception for autonomous driving \cite{gauerhof2018structuring}. Especially for such applications, it is crucial to understand the generalization performance of DNNs. While perfect generalization is difficult to achieve in practice, it is particularly important in situations where generalization issues are exploited on purpose. This is the case for adversarial examples \cite{goodfellow2014explaining}, \cite{kurakin2016adversarial}, \cite{papernot2016limitations}, \cite{carlini2017towards}, i.e. small perturbations of the input, that are often unrecognizable to humans, but lead to a misclassification by the network.\\
There are two ways to tackle the problem of adversarial examples. One is to use \textit{defense techniques}, which make DNNs robust against adversarial attacks by training. However, for each defense technique a new adversarial attack strategy can usually be found. To avoid this problem, other works build a method on top of the original network to \emph{detect} adversarial examples \cite{ma2018characterizing}, \cite{meng2017magnet}, \cite{liao2018defense}, \cite{xu2018feature}, \cite{ma2019nic}, \cite{lee2018simple}. Unfortunately, the current state-of-the-art detectors require either significantly more parameters or more runtime than the original network itself. This is critical for applications like autonomous driving where the available resources are limited.\\
We therefore propose \textbf{GraN}, a parameter-efficient method whose main component is the \textbf{Gra}dient of the network with respect to the weights and the current input-output combination, more precisely its \textbf{N}orm (see Figure \ref{Fig:MV}). GraN is based on the concept that misclassified examples have a larger gradient than correctly classified examples. It can be calculated in a manner similar to the way a training step is conducted. Our experiments on three different datasets show that GraN can detect misclassified examples faster and with less parameters, while performing comparable to state-of-the-art detectors.

\section{Related Work}
\label{sec:background}
Common adversarial detection methods can be split into three main categories \cite{ma2019nic}: denoisers, prediction inconsistency based approaches and metric based approaches. \\
\textbf{Denoisers} such as \textit{MagNet} \cite{meng2017magnet} and \textit{HGD} \cite{liao2018defense} are based on a preprocessing procedure, in which the image is reduced to its main features by an encoder-decoder method. The idea is to reduce noise and the added adversarial perturbation. Despite their good performance, denoisers can not be used in parameter-restricted approaches since an additional encoder-decoder network with many parameters is required \cite{liao2018defense}, \cite{ma2019nic}. \\
\textbf{Prediction inconsistency based approaches} use the idea that for misclassified images the classification output is more sensitive to small changes in the image. The most successful approach is based on \textit{feature squeezing}. The original image, an additional image with reduced color depth generated from the original image and a smoothed image are classified by the network. The probabilities for the three images are compared and the sensitiveness is used to decide whether the original image is misclassified \cite{xu2018feature}. Usually, only an insignificant number of parameters is used for such methods. However they do not perform well on some adversarial attacks \cite{xu2018feature}, \cite{ma2019nic}.\\
\textbf{Metric based approaches} use statistical methods to check whether the current input is behaving similar to normal inputs that have been investigated before. One of the best known is a detector based on the Local Intrinsic Dimensionality (\textit{LID}) computed on the layer wise activation space \cite{ma2018characterizing}. It is a weighted distance metric based on the k-nearest-neighbors of 100 randomly chosen samples from the training set. This detector was outperformed by a method using the layer wise \textit{Mahalanobis} distance to the closest class-conditional Gaussian distribution \cite{lee2018simple}. The most recent detector is based on Neural-network Invariant Checking (\textit{NIC}) \cite{ma2019nic}. Additionally to the layer-wise activation distribution a classification network consisting of one fully connected softmax layer is trained on each activation layer. The more the class probabilities differ from layer to layer, the more probable the current example is misclassified. \textit{LID}, \textit{Mahalanobis} and \textit{NIC} lead to good results but require a huge overhead of parameters or runtime in comparison to the actual classification network. In case of \textit{LID}, 100 images need to be saved for comparison in the activation spaces. For \textit{Mahalanobis} the mean and the covariance matrix for each class and often high-dimensional layer have to be stored and \textit{NIC} additionally requires a huge overhead of parameters due to the fully connected network on top of each activation layer.\\
Although some detectors show compelling performance, none of the existing detectors meet the computational complexity restrictions of autonomous driving.

\section{GraN: A Gradient-Norm Based Detector}
\label{sec:gran}

\begin{figure}[t]
	\small
	\centering
	\includegraphics[scale=0.18]{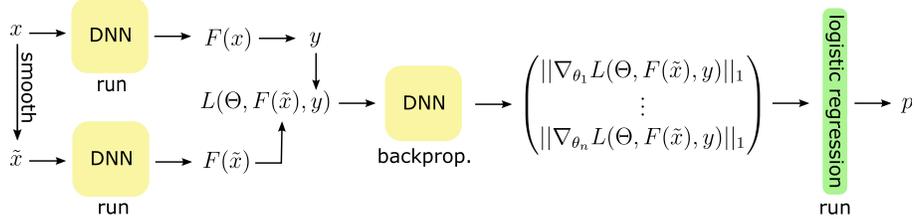}
	\caption{Overview of GraN.}\label{Fig:MV}
\end{figure}

In this section we first provide \textbf{GraN} as algorithm and secondly give information on the used strategies for the \textbf{indication of uncertainty}.\\
\textbf{GraN} is a detector predicting if a pre-trained DNN with weights $\Theta$ is misclassifying an input $x$. A flowchart of our approach is shown in Figure \ref{Fig:MV}. The method starts by running the DNN for the current input $x$. As usual for the classification, the index with the largest value of the output $F(x)$ is defined as the class prediction $y$ of the network. The same DNN is run for an input $\tilde{x}$, which is generated by smoothing $x$ with a Gaussian-kernel with standard deviation $s$. Both, the output $F(\tilde{x})$ on $\tilde{x}$ and the predicted class $y$ are fed into the loss function $L(\Theta,F(\tilde{x}),y)$. Then, the gradient with respect to the weights $\nabla_{\Theta} L(\Theta,F(\tilde{x}),y)$ is computed time-efficiently via backpropagation as in the training phase. In the next step the set of gradients $\nabla_{\Theta} L(\Theta,F(\tilde{x}),y)$ are transformed into a feature vector: For each layer $i \in \lbrace1,\dots,n\rbrace$ of the network, the gradient regarding the set of its weights $\theta_i$ is replaced by its $L_1$ norm
$$\nabla_{\Theta} L(\Theta,F(\tilde{x}),y) ={\begin{pmatrix}\nabla_{\theta_1} L(\Theta,F(\tilde{x}),y)\\\vdots \\\nabla_{\theta_n} L(\Theta,F(\tilde{x}),y)\end{pmatrix}}~~~\rightarrow~~~ {\begin{pmatrix}||\nabla_{\theta_1} L(\Theta,F(\tilde{x}),y)||_1\\\vdots \\||\nabla_{\theta_n} L(\Theta,F(\tilde{x}),y)||_1\end{pmatrix}}.$$
The resulting vector has length $n$ depending on the numbers of layers in the DNN. As in reference \cite{ma2018characterizing}, a logistic regression network with $n+1$ parameters is trained to predict the probability $p$ of the input $x$ to be misclassified.\\
The central idea and hence the main \textbf{indication of uncertainty} is the norm of the gradient of the loss function $\nabla_{\Theta} L(\Theta,F(x),y)$ computed for the network output $F(x)$ and the resulting class $y$. Mathematically, the gradient over a variable expresses the effect of small changes of that variable on the output. Therefore, the bigger the gradient, the bigger the corresponding uncertainty.\\
The difference between the network output $F(x)$ and the resulting one-hot class vector $y$ is one of the most simple indications of uncertainty. This evidence is covered by the gradient $\nabla_{\Theta} L(\Theta,F(x),y)$, since the difference directly builds the gradient over the last layer and represents the starting point for the backpropagation. Hence, a huge distance between $F(x)$ and $y$ directly increases all gradient values.\\
To further enlarge the gradient for misclassified examples, the smoothed input $\tilde{x}$ is used for the prediction $F(\tilde{x})$. This smoothing procedure is thought to remove the perturbations responsible for misclassification. Thereby we increase the difference between the class $y$ predicted for $x$ and the network output $F(\tilde{x})$ and hence the gradient $\nabla_{\Theta} L(\Theta,F(\tilde{x}),y)$.

\section{Evaluation}
\label{sec:evaluation}
We evaluate GraN on five adversarial attack methods and two additional misclassification tasks each applied on three datasets. To be able to directly compare the performance of GraN, we choose to implement LID as current state-of-the-art method with the least number of parameters as discussed in Section \ref{sec:background}.

\subsection{Experimental Setup}
All experiments are performed on the MNIST \cite{lecun1990handwritten}, CIFAR-10 \cite{krizhevsky2009learning} and SVHN \cite{netzer2011reading} datasets. All datasets consist of predefined test and training sets which we refer to as pre-train and pre-test sets, since for the detection additional test and training sets are required. For each dataset a DNN for the classification task is trained on the pre-training data and tested on the pre-test set. The DNNs are taken from reference \cite{ma2018characterizing}. For MNIST, a 5-layer Convolutional Neuronal Network (CNN) was used. It achieved an accuracy of 96.89\%. A 12-layer CNN was trained for CIFAR and an accuracy of 87.47\% was reached. For SVHN, a 6-layer CNN was trained and an accuracy of 90.29 \% was observed.\\
Similar to \cite{ma2019nic} and \cite{xu2018feature}, the \emph{adversarial set-up} is built from the correctly classified images from the pre-test set. For each of the images, five adversarial images are generated by the attack methods and settings reference \cite{ma2018characterizing} evaluated on: FGSM \cite{goodfellow2014explaining}, BIM-a, BIM-b \cite{kurakin2016adversarial}, JSMA \cite{papernot2016limitations} and CW \cite{carlini2017towards}. Only the adversarial examples that lead to misclassification are kept. Each attack method defines together with the correct classified images one dataset. Additionally, two set-ups are build that more closely resemble situations with real world sensor data. For the \emph{noisy set-up}, the correct classified images from the pre-test set are perturbed with Gaussian noise, such that one half of the noisy images is misclassified. For the \emph{wrong set-up}, the original complete pre-test dataset is taken including the misclassified examples. Each data set-up is split into a train (80\%) and a test (20\%) set such that each set contains an equal number of misclassified and correctly classified examples, excessive samples are deleted.\\
For GraN, the standard deviation for the Gaussian smoothing is set to $s=0.4$ for all datasets. The number of neighbors for LID was set as determined in reference \cite{ma2018characterizing}: $k=20$ for MNIST and CIFAR and $k=30$ for SVHN. For each set-up, the detectors are trained on the generated training set to predict the probability for an input to be misclassified and tested on the generated test set. The performance metric is the AUC ROC score $[$\%$]$. It is defined as the area under the receiver operating characteristic (ROC) curve, which plots the true positive rate over the false positive rate. The results are provided in Table \ref{tab:compare}.
\subsection{Discussion}

\begin{table}[t]
	\small
	\centering
	\begin{tabular}{|c |l| r r r r r r r| }
		\hline
		\multirow{2.3}{2mm}{D.} & \multirow{2.3}{2mm}{Met.} &
		\multicolumn{7}{c|}{Cause for misclassification}\\
		
		&   & FGSM & BIM-a & BIM-b & JSMA & CW  &Wrong & Noisy \\
		\hline
		
		\multirow{2.3}{*}{M}
		& LID     & \textbf{100.00} &  \textbf{99.70} & \textbf{99.99} & \textbf{99.25} & \textbf{99.87} &\textbf{97.35}  & \textbf{96.33}\\
		
		& GraN	& 99.51 &  98.46 &99.19 & 95.78 & 92.03 &93.25 &93.96 \\
		
		\hline
		\multirow{2.3}{*}{C}
		& LID       &\textbf{99.20} & 82.24 & \textbf{100.00} & 98.81 & 99.20  &  87.25  & 81.74 \\
		
		& GraN	& 98.82 & \textbf{84.70}  & 99.74  & \textbf{99.42} & \textbf{99.90} &\textbf{ 89.30}  &  \textbf{89.21}\\
		\hline
		\multirow{2.3}{*}{S}
		& LID      & \textbf{99.99} &87.26 & \textbf{99.98} &97.76 & 97.34 & 83.49 & 80.81 \\
		& GraN & 99.98  & \textbf{91.94} & 99.96 & \textbf{99.07} & \textbf{99.91}  & \textbf{91.86}  & \textbf{87.75}  \\
		\hline		
	\end{tabular}
	\caption{Performance of GraN (own) and LID \cite{ma2018characterizing} based on AUC ROC score $[$\%$]$ on different set-ups for MNIST (M), CIFAR-10 (C) and SVHN (S) dataset.}\label{tab:compare}
\end{table}
The AUC ROC performance for GraN is between 90\% and 100\% for almost all set-ups. Except for MNIST, GraN outperforms LID in five of seven set-ups. The slightly worse performance on MNIST can be lead back to the simplicity of the MNIST dataset. Even with many small changes the original number is still recognizable for humans, whereas the network is concentrating on a small number of features that in most cases are enough for classification. This makes it easy for adversarial attacks because small details can be easily changed and hence the gradient stays low. The more complex the dataset is, the more diverse feature information is saved within the network. Even if the main features are changed, there are several other features contradicting the false decision, which therefore leads to a higher gradient. In most application fields, the data is more complex than the simple MNIST dataset. This promises good performance of GraN on such.\\
We remove perturbations leading to misclassification irrespective if the input is adversarial or not, by means of the Gaussian low-pass filter as provided in Section \ref{sec:gran}. It was found empirically, that the usage of $F(\tilde{x})$ instead of $F(x)$ is leading to a performance improvement but is not essential for GraN's functionality. This perturbation removal procedure works very well in particular when the perturbations are similar to sensor noise. For autonomous driving applications this is a major use case. However, as our experiments show, it works as well for a larger class of adversarial perturbations.\\
The necessary resources for GraN and LID are compared using the example of the CIFAR dataset \cite{krizhevsky2009learning}: In addition to the parameters of the classification network, LID needs ($100\cdot 32\cdot 32\cdot 3 + 66=$) 307.266 parameters to save 100 images and 66 parameters for its logistic regression network. GraN only needs the 35 parameters for its logistic regression network and therefore 0.01\% from that of LID. Although there was no particular attention payed to runtime optimization, the computation time of GraN was 0.5\% from that of LID. There is the option to reduce the runtime of LID by directly saving the activations for each image, however this would lead to an even larger overhead of parameters for LID and the computation time of GraN would still be less than 50\% of that of LID.
\section{Conclusion and Future Work}
\label{sec:conclusion}
We introduced and evaluated GraN, a novel, parameter-efficient method that detects adversarial examples as well as other misclassified examples. It is able to distinguish between noisy examples that are correctly and incorrectly classified. The time- and parameter-efficient performance for various adversarial attacks, especially on the more complex data sets, make GraN a promising method for several real-world scenarios. The perturbation-removal approach used here is geared towards sensor-noise. Even though this seems rather specific, our experiments show that it works well for various adversarial attacks. In future work, we will assess the performance of other perturbation-removal procedures than Gaussian smoothing and investigate the effects of replacing the norm and logistic regression part with more powerful procedures.\\
\\
\textbf{Acknowledgment:} We thank Matthias Rath and Stephan Scheiderer for their valuable input.


\begin{footnotesize}

\bibliographystyle{unsrt}
\bibliography{literature}

\end{footnotesize}
\newpage


\end{document}